\documentclass[journal]{IEEEtran}
\usepackage[utf8]{inputenc}
\usepackage{amssymb}
\usepackage{graphicx}
\usepackage{booktabs}
\usepackage{multirow}
\usepackage{cite}
\usepackage{subfigure}
\usepackage{enumerate}
\usepackage[switch]{lineno} 
\usepackage{lipsum} 
\usepackage{colortbl}
\usepackage{hhline}
\usepackage{amsmath}
\usepackage{bm}
\usepackage{comment}
\usepackage{upgreek}
\usepackage{comment}
\usepackage{url}

\newcommand{\ce}[1]{{\color{blue} #1}}

\arrayrulecolor{black}

\newcommand{\GP}{\mathcal{GP}}

\ifCLASSOPTIONcompsoc
  \usepackage[caption=false,font=normalsize,labelfont=sf,textfont=sf]{subfig}
\else
  \usepackage[caption=false,font=footnotesize]{subfig}
\fi
\begin{document}

\title{Deep Feature Gaussian Processes for Single-Scene Aerosol Optical Depth Reconstruction}

\author{Shengjie~Liu,~\IEEEmembership{Student Member,~IEEE,}
        and~Lu~Zhang
\thanks{Manuscript received Apr 1, 2024. This work was supported by the Southern California Environmental Health Sciences Center (Grant No. P30ES007048). \emph{(Corresponding author: Lu Zhang.)}} 
\thanks{S. Liu is with the Spatial Sciences Institute, Dornsife College of Letters, Arts and Sciences, University of Southern California, Los Angeles, CA 90089, USA (e-mail: skrisliu@gmail.com).}
\thanks{L. Zhang is with the Division of Biostatistics, Department of Population and Public Health Sciences, Keck School of Medicine, University of Southern California, Los Angeles, CA  90033, USA (e-mail:  lzhang63@usc.edu).}
}

\markboth{Submitted to IEEE Geoscience and Remote Sensing Letters}%
{Submitted to IEEE Geoscience and Remote Sensing Letters}

\maketitle

\begin{abstract}
Remote sensing data provide a low-cost solution for large-scale monitoring of air pollution via the retrieval of aerosol optical depth (AOD), but is often limited by cloud contamination. Existing methods for AOD reconstruction rely on temporal information. However, for remote sensing data at high spatial resolution, multi-temporal observations are often unavailable. In this letter, we take advantage of deep representation learning from convolutional neural networks and propose Deep Feature Gaussian Processes (DFGP) for single-scene AOD reconstruction. By using deep learning, we transform the variables to a feature space with better explainable power. By using Gaussian processes, we explicitly consider the correlation between observed AOD and missing AOD in spatial and feature domains. Experiments on two AOD datasets with real-world cloud patterns showed that the proposed method outperformed deep CNN and random forest, achieving R$^2$ of 0.7431 on MODIS AOD and R$^2$ of 0.9211 on EMIT AOD, compared to deep CNN's R$^2$ of 0.6507 and R$^2$ of 0.8619.  The proposed methods increased R$^2$ by over 0.35 compared to the popular random forest in AOD reconstruction. The data and code used in this study are available at \url{https://skrisliu.com/dfgp}. 
\end{abstract}


\IEEEpeerreviewmaketitle

\ce{This is the preprint version. For the final version, go to IEEE Xplore: \url{https://doi.org/10.1109/LGRS.2024.3398689}.}

\section{Introduction}

As a major environmental exposure affecting people's health, air pollution is estimated to cost around 5.5 million premature deaths worldwide annually \cite{lelieveld2020loss}. Air pollution monitoring relies on ground stations that often have sparse geographical distributions, especially in densely populated urban areas due to the expensive cost to set up and maintain stations in prime city locations \cite{zhu2017extended}. Spaceborne satellite data provide an alternative low-cost solution for spatially-continuous monitoring of air pollution via their retrieval of aerosol optical depth (AOD) -- a physical property of the atmosphere that is highly related to air pollution \cite{li2005retrieval}. Although many retrieval algorithms have been developed \cite{fan2023satellite}, since on average 67\% of the Earth's surface is covered by clouds, satellite-retrieved AOD data are often contaminated and incomplete spatially \cite{king2013spatial}. 

To tackle the issue of incomplete AOD data due to cloud contamination, many methods have been proposed by using multi-temporal observations. For example,  Yang and Hu \cite{yang2018filling} used spatiotemporal kriging to gap fill MODIS AOD, increasing data completeness from 14\% to 68\%. Goldberg et al. \cite{goldberg2019using} used the inverse distance weighting to gap fill MODIS AOD to achieve seamless mapping of PM2.5 concentration.  Li et al. \cite{li2019comparative} proposed to use autoregressive integrated moving averaging to interpolate MODIS AOD for studying seasonal patterns. All the above methods can produce reasonably good results in AOD reconstruction in a spatiotemporal fashion. However, for satellite data at high spatial resolution, such as Landsat and other sensors onboard low Earth orbit satellites (e.g., the EMIT [Earth surface Mineral dust source InvesTigation] sensor installed on the International Space Station), multitemporal observations are often unavailable. This is due to their low Earth orbit nature and long re-visiting time, e.g., 16 days for Landsat. The lack of multitemporal observations is also and especially true for AOD estimation using above-cloud airborne remote sensing data when only one snapshot is available \cite{leblanc2020above}. 

For single-scene AOD data reconstruction, random forest has been proven effective with the usage of explanatory variables including normalized difference vegetation index (NDVI), digital elevation model, and land surface temperature \cite{li2021variability}. Although deep learning and convolutional neural networks (CNNs) have achieved better performance than random forest in many remote sensing tasks including hyperspectral image classification \cite{liu2020few}, PolSAR image classification \cite{bi2019active} and soil moisture reconstruction \cite{fang2018value}, their usage in AOD data reconstruction is limited, mostly limited to spatiotemporal settings \cite{lops2021application}. Deep CNNs can extract and transform variable features via their convolutional structures, nonlinear activation functions such as the rectified linear unit (ReLU) and sigmoid, thereby achieving prime performance \cite{zhang2016deep}. 

Unlike multitemporal AOD reconstruction, there is no temporal information in single-scene AOD reconstruction. Therefore, explainable variables and spatial information become crucial in accurately estimating AOD values. Gaussian processes (GPs) provide a modeling tool to fully consider spatial information based on the physical proximity of data samples \cite{camps2016survey,zhang2019practical}. Nevertheless, the implementation of GPs faces significant challenges due to the computational cost and storage requirements, which scale cubically and quadratically with the size of the training data, respectively. This escalation quickly becomes prohibitive as the data scale increases, impeding its application in AOD satellite data modeling. Moreover, GP-based regression models typically assume a constant mean or linear regression of predictors, a simplification that further constrains their effectiveness when dealing with the complexities inherent in satellite data.

In the era of deep learning, deep representation learning to achieve nonlinear feature transformation is key to the unprecedented performance of deep CNNs. On the other hand, GPs provide a probabilistic approach to model intrinsic correlation across observations, enabling a more nuanced understanding of data uncertainty and variability. Thus, in this letter, we propose to use CNNs to extract the nonlinear deep features and to use GPs on the transformed deep features to harness the rich, probabilistic insights of the global information, thereby achieving single-scene AOD reconstruction. The proposed method is therefore named deep feature Gaussian processes (DFGP). To further facilitate the inference of DFGP, we propose to use a scalable variational approximation in GPyTorch that allows us to handle millions of data points with GPU-accelerated computing \cite{hensman2015scalable, gardner2018gpytorch}. Although there are also other approaches to achieve GPs, including using deep CNNs as an approximation \cite{lee2017deep}, using scalable GPs allows fast computation. To sum up, the major contributions of this letter are as follows. 
\begin{itemize}
    \item We propose a novel method for single-scene AOD reconstruction named deep feature Gaussian processes (DFGP) that maintains the advantage of GPs to fully consider spatial information and also has the deep representation learning power. 
    \item We use a scalable variational approximation of GPs in GPyTorch so that the proposed method can easily handle millions of data points, rather than $\sim$10,000 data points in conventional GPs. 
    \item The proposed method can offer confidence intervals and uncertainty estimates under Gaussian assumption. 
\end{itemize}

\section{Methodology}
\subsection{Deep Convolutional Neural Networks}
The advantage of deep CNNs is that they can extract and transform features nonlinearly. For feature extraction, the convolution operations allow the model to explore local spatial features within a $k\times k$ window. For each pixel, we have
\begin{equation}
    x^{\phi_1}[i,j] = Conv_{\phi_1} (x[i,j]) =  \sum_{q=-k/2}^{k/2} \sum_{p=-k/2}^{k/2} \omega_{p,q}^{\phi_1} \cdot x[i+p,j+q],
\end{equation}
where $\phi_1$ denote a feature transformation (here via convolution), $i,j$ is the index of the data point within a two-dimensional image, $\omega$ are weights of the convolution, $x[i,j]$ is the original feature at location $[i,j]$, and $x^{\phi_1}[i,j]$ is the transformed feature at location $[i,j]$. Also, as multiple convolution operations stack together, the receptive field of deep CNNs increases. Additionally, nonlinear activation functions such as batch normalization and ReLU help facilitate the feature transformation process \cite{ioffe2015batch,nair2010rectified}. In the training process, for each training batch $B$ where the sample index is $i$  and $x^{\phi_1}_i \in B$, batch normalization transforms features from $x^{\phi_1}_i$ to $x^{\phi_2}_i$ via the following transformation:
\begin{equation}
    x^{\phi_2}_i = \frac{x^{\phi_1}_i-\mu_{B}}{\sqrt{\sigma_B^2+ \epsilon} }
\end{equation}
where $\mu_B = \frac{1}{N_{B}} \sum_{i}^{N_B} x^{\phi_1}_{i}$ is the mean of feature values $x^{\phi_1}$ in this batch $B$, and $\sigma^2_B = \frac{1}{N_{B}} \sum_{i}^{N_B} (x^{\phi_1}_i - \mu_B)^2 $ is the variance. In this letter, the wide contextual residual network (WCRN) \cite{liu2020active} with one residual block (one multi-level convolutional layer, two batch normalization layers, two ReLU function, and two convolutional layers) was adapted due to its efficiency and good performance even with limited training samples. Note that the proposed method should work with any deep CNNs (we added HResNet \cite{liu2020few} to demonstrate this flexibility). The original WCRN was designed for classification; we replaced the multi-class fully-connected layer with softmax function to a fully connected layer with sigmoid function for AOD reconstruction. This revision was made after we obtained the final transformed features $x^\phi$ via feature extraction in the network
\begin{equation}
\label{eq:network}
    x^{\phi} = \phi(x). 
\end{equation}
And the entire network can be optimized by minimizing the $\ell_1$ loss: 
\begin{equation}
  \ell_1 = \sum^{N_{\mathbb{B}}}_{i} \left| y_i - \frac{1}{1+\exp{( - w_{fc} \cdot x^{\phi}_i)} } \right|,
\end{equation}
where $w_{fc}$ are the weights of the fully-connected layer, $y_i$ is training label,  and $N_{\mathbb{B}}$ is the number of all training data. This optimization process is via mini-batch gradient descent. 

\subsection{Gaussian Processes}
GPs explicitly consider global information from the spatial or feature space, or both. Let $f(s)$ denote the spatial process of the outcome of interest measured at location $s$ in study domain $\cal{D}$. Assume that $f(s)$ follows a GP with mean function $m(s)$ and covariance function $k(\cdot, \cdot)$
\begin{equation}
\label{eq:gp}
f(s) \sim \GP (m(s), k(\cdot, \cdot))\;,
\end{equation}
where $k(s, s’)$ defines the covariance between $f(s)$ and $f(s’)$.
Given a training set $ \mathcal{S}_{tr} = \{s_i, \mathbf{x}_i, y_i\}_{i=1}^{N} $, where $N$ is the number of training samples, $x_i$ is the predictors at the $i$-th location $s_i$, and $y_i$ is the outcome of interest. Denote $\mathbf{s} = [s_1, ..., s_N]^\top$,  $\mathbf{X}=[ \mathbf{x}_1: ...: \mathbf{x}_N]^\top$, $\mathbf{y} = [y_1, ..., y_N]^\top $, $\mathbf{K} = [\mathbf{k}_1 : \mathbf{k}_i : ... : \mathbf{k}_N]^\top  $ where $\mathbf{k}_{i} = k(s_i,\mathbf{s})$; $f(\mathbf{s})$ follows a multivariate normal distribution
\begin{equation}
\label{eq:gp2}
f(\mathbf{s}) \sim \mathcal{N}_N (m(\mathbf{s}), \mathbf{K} )\;.
\end{equation}
The mean function $m(\mathbf{\cdot})$ in GPs depends on $\mathbf{X}$ and is customarily modeled using linear regression, and the covariance function is often the radial basis function (RBF) or Mat\'ern kernels. We can find a point estimator of the parameters $\Theta$ in linear regression and in the kernel via maximum a posteriori (MAP) from the log-likelihood function \cite{rasmussen2006gaussian}  
\begin{align}
  &\begin{aligned}
  \log p (\mathbf{y} | \mathbf{X}, \mathbf{K}, \Theta )  =   &  - \frac{1}{2} (\mathbf{y}-m(\mathbf{X})  )^\top \mathbf{K}^{-1} (\mathbf{y} - m(\mathbf{X}) )  \\
  & - \frac{1}{2} \log |\mathbf{K}| - \frac{n}{2} \log (2\pi)\;.
  \end{aligned}
\end{align}
Given a set of unobserved data $\mathcal{S}_* = \{ s_{i*}, \mathbf{x}_{i*} \}_{i=1}^{N_*} $, where $N_*$ is the number of unobserved locations $\mathbf{s}_*$ and $\mathbf{X}_* = [ \mathbf{x}_{1*}; ...: \mathbf{x}_{N_*}]^\top $ records their predictors, under the GP assumption, the predictive distribution of $f(\mathbf{s}_*)$ for unobserved locations $\mathbf{s}_*$  follows $ f(\mathbf{s}_*) \sim \mathcal{N}_{N_*} (\boldsymbol{\mu}_*,\mathbf{K}_*)$, where 
\begin{equation}\label{eq:GP_par}
\begin{aligned}
    \boldsymbol{\mu}_* &= m(\mathbf{X}_*) + k(\mathbf{s}_*,\mathbf{s}) \mathbf{K}^{-1}(\mathbf{y} - m(\mathbf{X} ) )\;,\\
    \mathbf{K}_* &= k(\mathbf{s}_*,\mathbf{s}_*) - k(\mathbf{s}_*,\mathbf{s})\mathbf{K}^{-1} k(\mathbf{s}_,\mathbf{s}_*)\;.
\end{aligned}
\end{equation}
We used the Mat\'ern kernel to model dependence: 
\begin{equation}
    k_{M} \left(\mathbf{x}, \mathbf{x}' \right) = \frac{2^{1-\nu}}{\Gamma(\nu)}\left(\frac{\sqrt{2 \nu} d }{\ell}\right)^{\nu} K_{\nu}\left(\frac{\sqrt{2 \nu} d}{\ell}\right)  \;,
\end{equation}
where $d = \| \mathbf{x} - \mathbf{x}' \|$ is the distance between $\mathbf{x}$ and $\mathbf{x}'$, $\Gamma$ is the gamma function, $K_{\nu}$ is the modified Bessel function, and $\ell$ and $\nu$ are hyperparameters. We set the length scale $\ell$ constraint to values corresponding to 3-84 km for MODIS (1 km) and to 0.25-13 km for EMIT (60 m) to ensure its physical meaning in DFGP$_s$. This value is relaxed to half range in DFGP. We compared the Mat\'ern and RBF kernels in sensitivity analysis later.

\subsection{Deep Feature Gaussian Processes}
A limitation of conventional GPs lies in their mean function, which is often simple linear regression. Given the complexity of satellite data, simple linear regression has been proven ineffective. Our proposed method replaces the variable from its original feature space $\mathbf{x}$ to a transformed feature space $\mathbf{x}_\phi$ via a CNN (see Eq. \ref{eq:network}), and then Eq. \ref{eq:gp} can be rewritten as 
\begin{equation}
f(s) \sim \GP(\tilde{m}(s), \tilde{k}(\cdot, \cdot))
\end{equation}
 where $\tilde{m}(s)$ depends on $\mathbf{x}_\phi (s)$, the predictors at location $s$ in the transformed feature space $\cal{X}_\phi$. The covariance function $\tilde{k}(x_\phi(s), x'_\phi(s'))$ defines the covariance between $f(s)$ and $f(s')$ based on their distance in the transformed feature space.
This feature transformation can also be viewed as using pre-trained models to obtain better features, and therefore the pre-trained model $\phi(\cdot)$ can be extended to any CNN that has the same input structure but not necessary the same output. In addition to the standard proposed method DFGP, we propose its variant DFGP$_s$ where the covariance function $k(s, s')$ is the spatial covariance between $s$ and $s'$:
\begin{equation}
    f(s) \sim \GP\left(\tilde{m}(s),k\left(s,s'\right)\right). 
\end{equation}
To facilitate the training on GPU, we used GPyTorch with a variational approximation of GP distribution obtained via variational evidence lower bound (ELBO) for experiments related to GPs \cite{hensman2015scalable,gardner2018gpytorch}.

\section{Experiments}

\subsection{Data}

\begin{figure}[!t]
\centering
\includegraphics[width=0.49\textwidth]{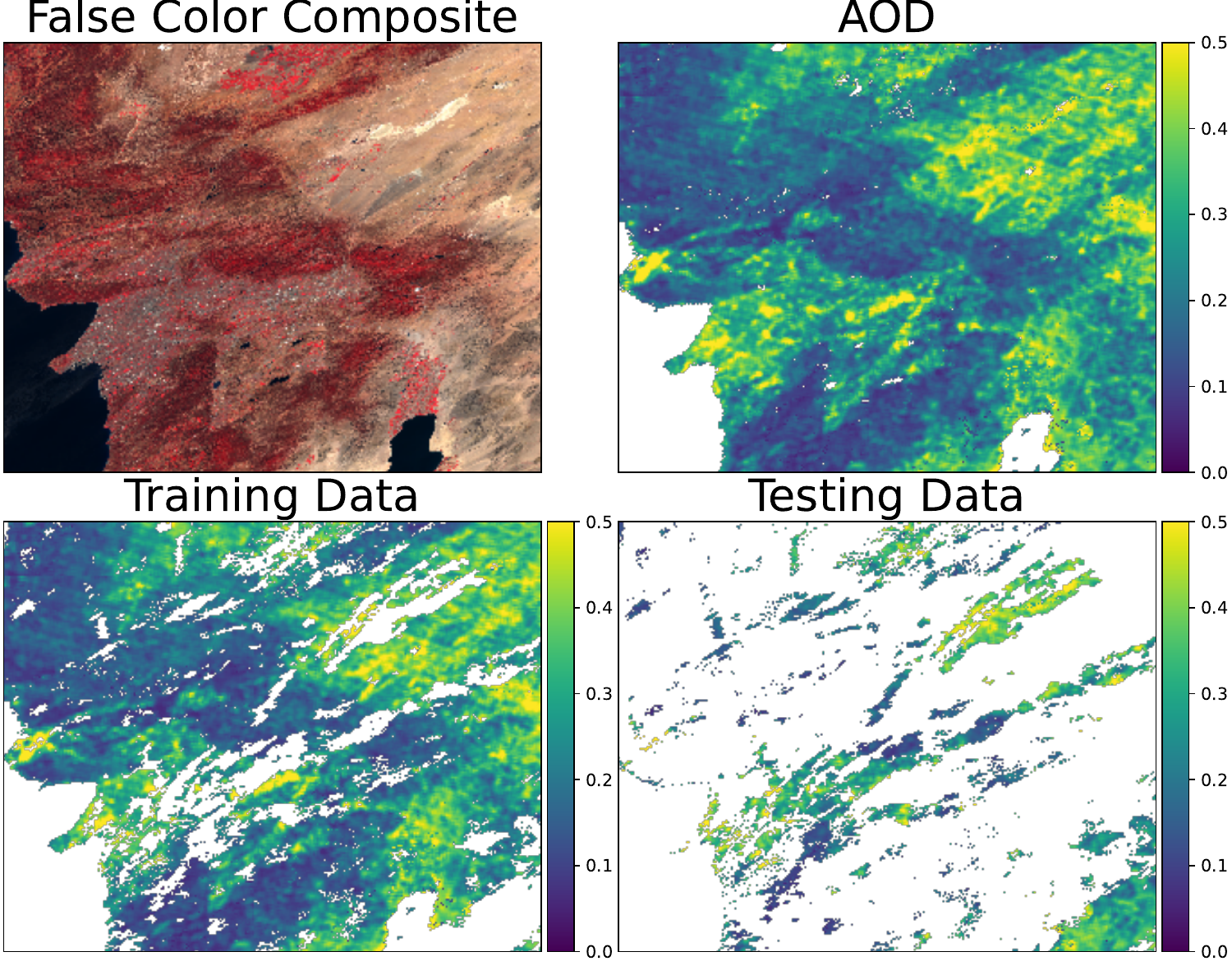} 
\caption{MODIS AOD, with 1 km spatial resolution.}
\label{fig:im1}
\end{figure}

Experiments were conducted on two datasets from two satellites. The first dataset is from 1-km MODIS AOD products (MCD19A2) over Los Angeles. The green band AOD from an almost cloud-free scene (2018-09-13) was used, with 65,793 valid samples out of a 300$\times$240. A cloudy date (2018-08-24) was used as the real-cloud pattern, leading to 53,540 training and 12,253 testing pixels. We collected 11 features for prediction, including 8 Landsat-8 spectral bands (median values in 2018), enhanced vegetation index (EVI) from MODIS MOD13A2, impervious surface percentage from USGS National Land Cover Database (NLCD) 2018, and road network density from OpenStreetMap.

\begin{figure}[!t]
\centering
\includegraphics[width=0.49\textwidth]{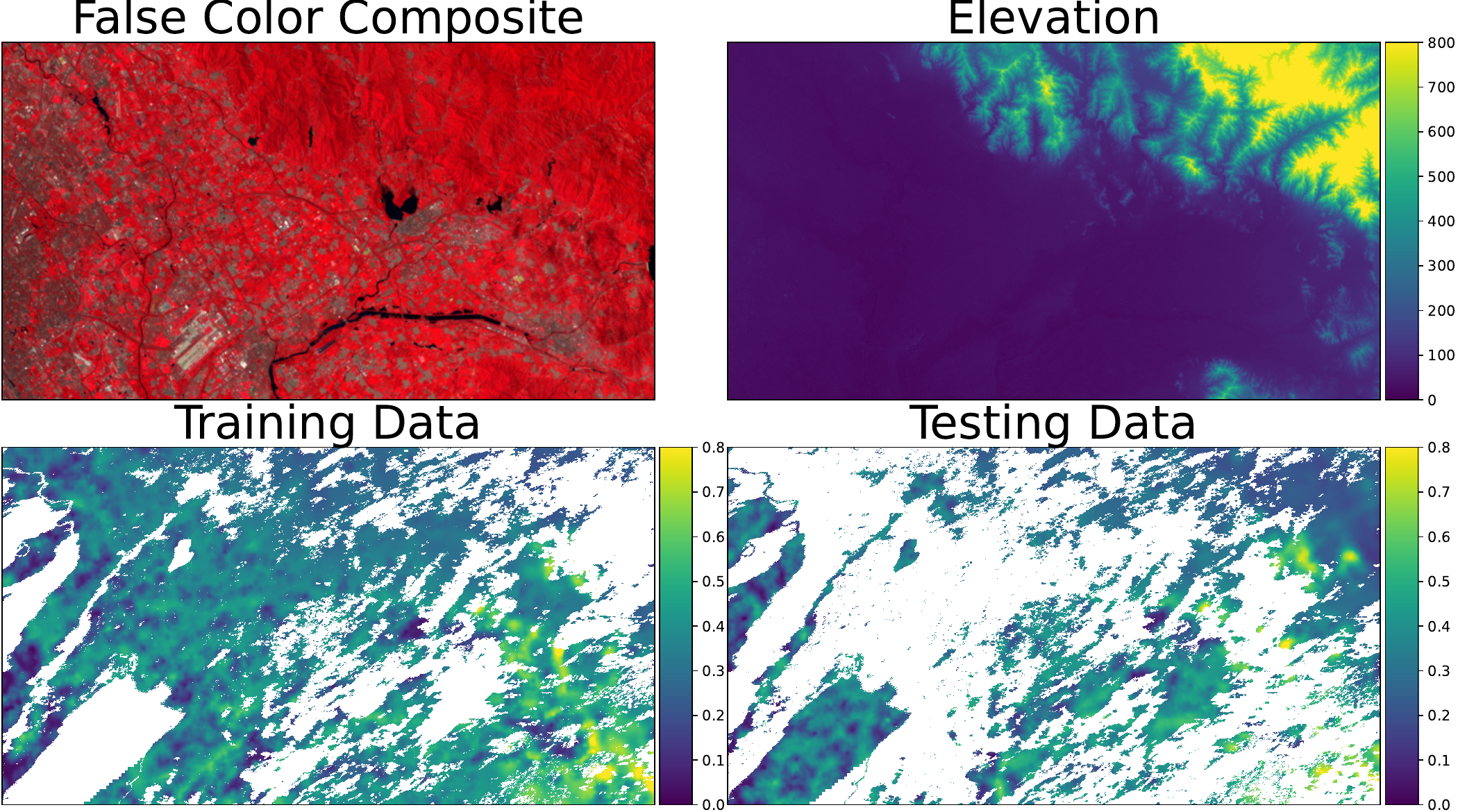}
\caption{EMIT AOD, with 60 m spatial resolution.}
\label{fig:im2}
\end{figure}

The second dataset is from the Earth Surface Mineral Dust Source Investigation (EMIT) instrument onboard the International Space Station (ISS), with a spatial resolution of 60 m \cite{green2020earth}. While the ISS orbits the Earth about every 90 minutes, the revisit time of a specific location depends. The sensor captures signals from 380 to 2,500 nm with 7.4 nm spectral sampling (285 bands in total) \cite{green2023performance}. The sensor was launched in July 2022. We used the  EMITL2ARFL data products from USGS. A cloud-free scene from 2023-08-19 over Beijing was used with a simulated real-world cloud pattern from MODIS. The post-processed dataset has a size of 570$\times$1,040 with 367,189 training pixels and 225,611 testing pixels. All 285 EMIT spectral bands from 380 to 2,500 nm are used as features in the experiments.

\subsection{Experimental Setup }
We compared the proposed method with six competitors: random forest (RF), linear regression (LR), conventional spatial GP (GP$_s$),  conventional GP (GP), and two baseline CNNs (WCRN \cite{liu2020active}, HResNet \cite{liu2020few}). For experiments on deep features, we added spatial variables (x, y) to construct covariance and also as features.

The CNN was trained on PyTorch with the Adam optimizer, with a batch size of 32 using a learning rate of 0.01 for 100 epochs and then of 0.001 for 50 epochs. The deep features were fed into the GP model on GPyTorch. We chose 500 training samples as the inducing points. The GP model was trained with the Adam optimizer with a batch size of 1024 using a learning rate of 0.01 for 300 epochs and then of 0.001 for additional 100 epochs. Linear regression and random forest were conducted using scikit-learn \cite{pedregosa2011scikit}. The number of trees was set as 500 in random forest. 
We report root mean squared error (RMSE), mean absolute error (MAE), and coefficient of determination (R$^2$) for evaluation. We conducted 10 random runs and reported the mean and standard deviation for algorithms using gradient descent.

\subsection{Results}
Table \ref{tab:table1} shows the results on MODIS AOD. DFGP$_s$ achieved the best result (R$^2$=0.7431), followed by DFGP (R$^2$=0.7048), compared to WCRN (R$^2$=0.6507). DFGP$_s$ has a better performance due to a more scattered cloud pattern, enabling the modeling of spatial dependence. Similar performance was achieved with deep features from HResNet, showing the generalization of the proposed methods. 

\begin{table}[htbp]
\begin{center}
\caption{Results on MODIS AOD. Standard deviation in brackets for methods using gradient descent. The best result is highlighted in \textbf{bold}. }
\scalebox{0.98}[0.98]{
\begin{tabular}{ |c|c|c|c|c| } 
 \hline
 Feature & Method & RMSE $\downarrow$ & MAE $\downarrow$ & R$^2$ $\uparrow$ \\
 \hline
 Original  & RF &  .0867 & .0639  &  .3866 \\
  & LR & .0784  & .0569  &  .4990 \\
 & GP$_s$ & .0636 \scriptsize{(.0024)} & .0463 \scriptsize{(.0020)} & .6694 \scriptsize{(.0053)} \\
 & GP & .0663 \scriptsize{(.0021)} & .0482 \scriptsize{(.0018)} & .6416 \scriptsize{(.0092)} \\
  \hline
 Deep  & WCRN & .0654 \scriptsize{(.0010)} & .0470 \scriptsize{(.0008)} &  .6507 \scriptsize{(.0032)} \\
 (WCRN) & DFGP$_s$ & \textbf{.0561} \scriptsize{(.0012)} &  \textbf{.0402} \scriptsize{(.0012)} & \textbf{.7431} \scriptsize{(.0029)} \\
 & DFGP & .0601 \scriptsize{(.0010)}  & .0432 \scriptsize{(.0009)}  & .7048 \scriptsize{(.0023)} \\
 \hline
 Deep  & HResNet & .0649 \scriptsize{(.0025)} & .0460 \scriptsize{(.0020)} & .6565 \scriptsize{(.0036)} \\
 (HResNet) & DFGP$_s$ & .0562 \scriptsize{(.0007)}  & .0402 \scriptsize{(.0005)}  & .7423 \scriptsize{(.0059)} \\
 & DFGP &  .0594 \scriptsize{(.0008)}  &  .0431 \scriptsize{(.0009)} & .7114 \scriptsize{(.0037)} \\
 \hline
\end{tabular}} 
\label{tab:table1}
\end{center}
\end{table}

Table \ref{tab:table2} shows the results on EMIT AOD. DFGP increased R$^2$ from WCRN's 0.8619 to 0.9211 and from HResNet's 0.8669 to 0.9228. DFGP has a better performance than DFGP$_s$, as the result of a more concentrated cloud pattern.

\begin{table}[htbp]
\begin{center}
\caption{Results on EMIT AOD. Standard deviation in brackets for methods using gradient descent. The best result is highlighted in \textbf{bold}. }
\scalebox{0.98}[0.98]{
\begin{tabular}{ |c|c|c|c|c| } 
 \hline
 Feature & Method & RMSE $\downarrow$ & MAE $\downarrow$ & R$^2$ $\uparrow$ \\
 \hline
 Original  & RF & .0590  & .0442  &  .7066 \\
  & LR & .0515 & .0353 &  .7767  \\
 & GP$_s$ & .0445 \scriptsize{(.0003)} & .0275 \scriptsize{(.0002)} & .8330 \scriptsize{(.0015)}  \\
 & GP & .0398 \scriptsize{(.0002)} & .0243 \scriptsize{(.0001)} &  .8666 \scriptsize{(.0008)} \\
  \hline
 Deep  & WCRN &  .0405 \scriptsize{(.0023)}  & .0186 \scriptsize{(.0012)}  & .8619 \scriptsize{(.0022)}  \\
 (WCRN) & DFGP$_s$ & .0340 \scriptsize{(.0016)} & .0168 \scriptsize{(.0013)}  & .9027 \scriptsize{(.0010)}   \\
 & DFGP & .0306 \scriptsize{(.0015)}  & .0163  \scriptsize{(.0014)}  & .9211 \scriptsize{(.0007)}  \\
  \hline
 Deep  & HResNet & .0397 \scriptsize{(.0013)}  & .0186 \scriptsize{(.0011)}  &  .8669 \scriptsize{(.0038)} \\
 (HResNet) & DFGP$_s$ & .0316 \scriptsize{(.0002)}  &  .0165 \scriptsize{(.0004)}  &   .9156 \scriptsize{(.0009)}  \\
 & DFGP & \textbf{.0304} \scriptsize{(.0004)} & \textbf{.0160} \scriptsize{(.0004)}  &  \textbf{.9228} \scriptsize{(.0027)} \\
 \hline
\end{tabular}}
\label{tab:table2}
\end{center}
\end{table}

\begin{figure}[!t]
\centering
\includegraphics[width=0.49\textwidth]{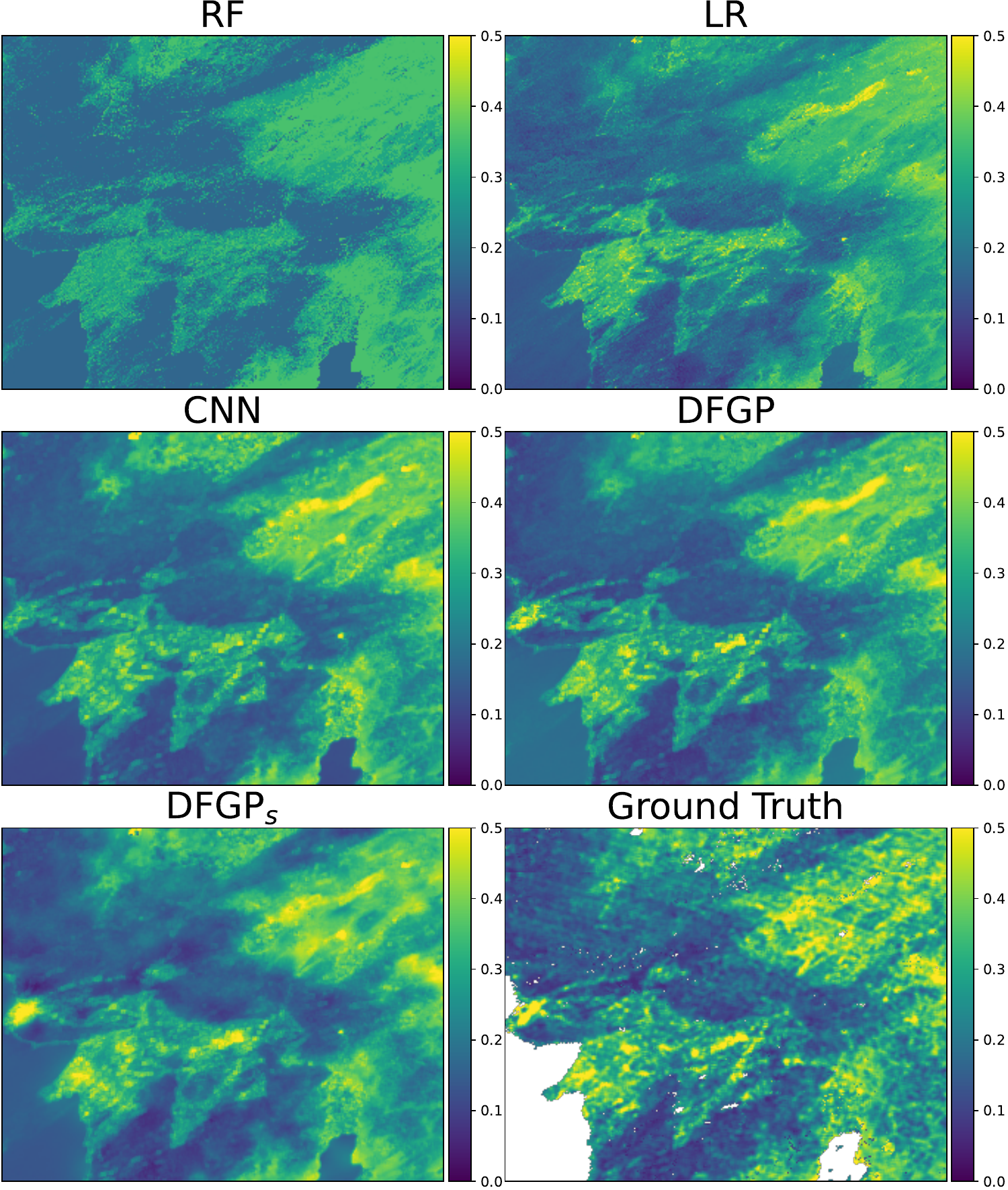} 
\caption{Reconstruction results, along with the ground truth.}
\label{fig:predict}
\end{figure}

Fig. \ref{fig:predict} shows the reconstruction results alongside with the ground truth. Reconstruction via classic methods such as RF and LR tend to be over-smoothed. The best reconstruction via DFGP$_s$ successfully captured the details in high-value regions.

\subsection{Sensitivity Analysis}

\begin{figure}[!t]
\centering
\includegraphics[width=0.24\textwidth]{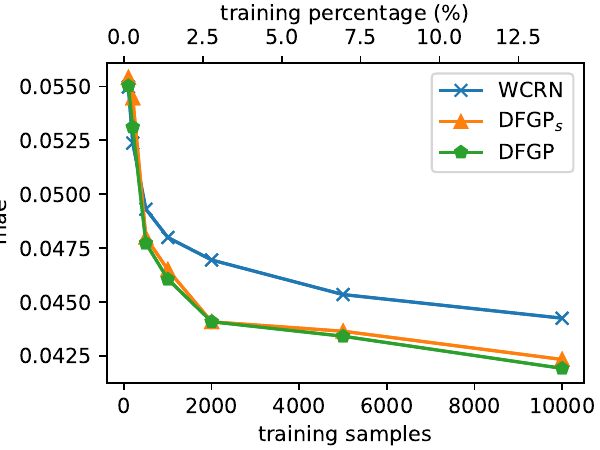} 
\includegraphics[width=0.24\textwidth]{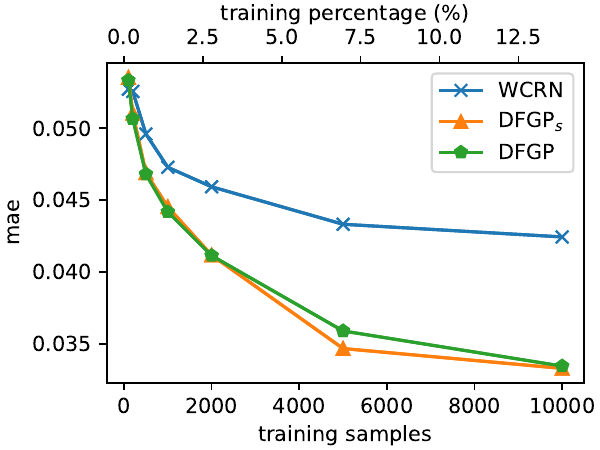}  \\
\includegraphics[width=0.24\textwidth]{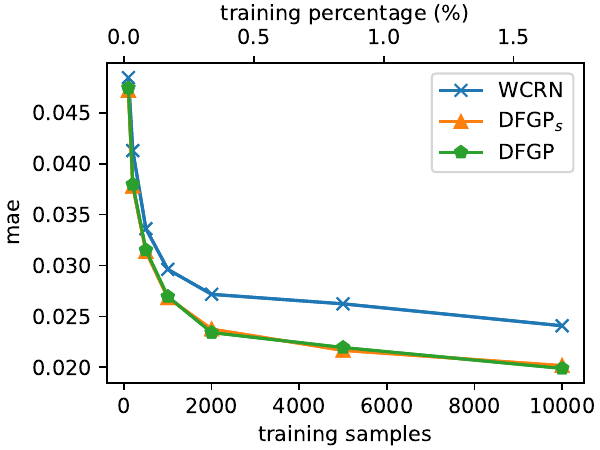} 
\includegraphics[width=0.24\textwidth]{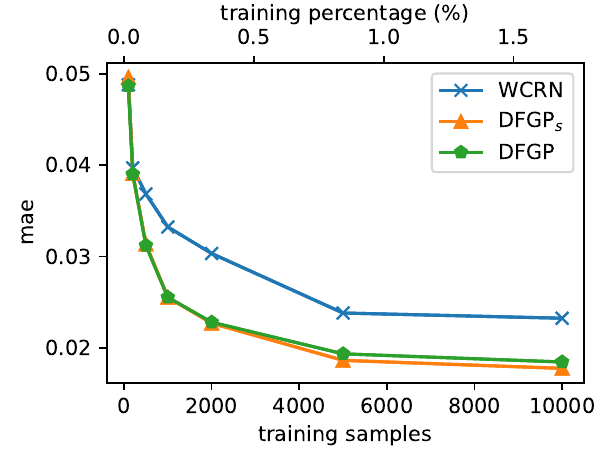} 
\caption{Mean absolute error (MAE) as a function of the number of training samples. \textit{(1st row: MODIS-Grid, MODIS-Random; 2nd row: EMIT-Grid, EMIT-Random.)}} 
\label{fig:sensitivity}
\end{figure}

Fig. \ref{fig:sensitivity} shows MAE as a function of training samples for both datasets. Samples were collected in two scenarios. For the Grid scenario, we divided the image into 10$\times$10 patches, and training pixels are only retrieved from 20\% of the patches to ensure no overlapping. The Grid scenario is more challenging and realistic. In the Random scenario, samples are randomly retrieved. All algorithms have better performance with random sample scenarios. With limited training samples, the proposed methods have similar performance with CNN due to the lack of data points to build the spatial/global dependence. With more samples ($>$500), regardless of datasets or scenarios, DFGP and DFGP$_s$ improved over CNN with lower MAE metrics. 

Table \ref{tab:table3} compares the performance of RBF and Mat\'ern kernels. The Mat\'ern kernel generally has a better performance. The RBF kernel has similar performance to Mat\'ern in DFGP but not in DFGP$_s$. The Mat\'ern kernel is less smooth due to its usage of 1st-order distance metric, compared to RBF's 2nd order distance metric. As a result, DFGP$_s$ with Mat\'ern kernel can better capture sharp changes in high-value regions (a type of spatial correlation).

\begin{table}[htbp]
\begin{center}
\caption{Comparison of RBF and Mat\'ern kernels on MODIS data}
\scalebox{0.98}[0.98]{
\begin{tabular}{ |c|c|c|c|c| } 
 \hline
 Method & Kernel & RMSE $\downarrow$ & MAE $\downarrow$ & R$^2$ $\uparrow$ \\
 \hline
 DFGP  & RBF & .0603 \scriptsize{(.0011)}  & .0435 \scriptsize{(.0009)}  & .7033 \scriptsize{(.0020)} \\
 & Mat\'ern &  .0601 \scriptsize{(.0010)}  & .0432 \scriptsize{(.0009)}  & .7048 \scriptsize{(.0023)} \\
  \hline
 DFGP$_s$ & RBF & .0569 \scriptsize{(.0009)} & .0410 \scriptsize{(.0008)} &  .7352 \scriptsize{(.0021)} \\
 & Mat\'ern &  .0561 \scriptsize{(.0012)}   &  .0402 \scriptsize{(.0012)}  &  .7431 \scriptsize{(.0029)} \\
 \hline
\end{tabular} }
\label{tab:table3}
\end{center}
\end{table}

\section{Conclusion}
In this letter, we proposed deep feature Gaussian processes (DFGP) for single-scene aerosol optical depth (AOD) reconstruction. The proposed method can take advantage from both the deep representation learning of CNNs and the probabilistic insights of the global information of GPs. Results on two datasets from two different sensors showed that the proposed DFGP and its variant DFGP$_s$ outperformed the popular method random forest and the state-of-the-art CNN. GPs are a modeling tool that offers many attractive qualities. It should be noted that the current implementation of DFGP is empirical Bayesian that tends to underestimate uncertainty, resulting in narrower credible intervals. For a more precise uncertainty quantification, extensive Markov chain Monte Carlo (MCMC) is one possible solution for full Bayesian sampling.

\section{Acknowledgements}
The authors would like to thank NASA, USGS, and OpenStreetMap for making their data publicly available. The authors would also like to thank the Editor, Associate Editor, and reviewers for their helpful comments and suggestions that improved this study.

\tiny{
\ifCLASSOPTIONcaptionsoff
  \newpage	
\fi
\bibliographystyle{IEEEtran}
\bibliography{strings}
}

\end{document}